\documentclass{article}
\usepackage{xcolor}

\usepackage[english]{babel}
\usepackage{multirow}
\usepackage{float}
\usepackage{xcolor}

\usepackage[letterpaper,top=2cm,bottom=2cm,left=3cm,right=3cm,marginparwidth=1.75cm]{geometry}
\usepackage{CJK}

\usepackage{authblk}
\usepackage{amsmath}
\usepackage{graphicx}
\usepackage[colorlinks=true, allcolors=blue]{hyperref}
\usepackage{multicol}
\title{\textbf {Trust \& Safety of LLMs and LLMs in Trust \& Safety}}

\author{Doohee You $^{1}$\footnote{Corresponding author at Google: doohee-at-google.com. \\      Second author at Google: danchon-at-google.com \\ Disclaimer: 1 = Google. The findings, interpretations, and conclusions expressed in this paper are entirely those of the author(s) and do not necessarily reflect the views of the Google or its affiliated entities.},   Dan Chon  $^{1}$}

\begin{document}
\maketitle

\begin{abstract}

In recent years, Large Language Models (LLMs) have garnered considerable attention for their remarkable abilities in natural language processing tasks. However, their widespread adoption has raised concerns pertaining to trust and safety. This systematic review investigates the current research landscape on trust and safety in LLMs, with a particular focus on the novel application of LLMs within the field of Trust and Safety itself. We delve into the complexities of utilizing LLMs in domains where maintaining trust and safety is paramount, offering a consolidated perspective on this emerging trend.\

By synthesizing findings from various studies, we identify key challenges and potential solutions, aiming to benefit researchers and practitioners seeking to understand the nuanced interplay between LLMs and Trust and Safety. 

This review provides insights on best practices for using LLMs in Trust and Safety, and explores emerging risks such as prompt injection and jailbreak attacks. Ultimately, this study contributes to a deeper understanding of how LLMs can be effectively and responsibly utilized to enhance trust and safety in the digital realm.

\end{abstract}

\newpage
\tableofcontents

\newpage
\newpage
\begin{multicols}{2}
\section{Introduction}

LLMs, powered by deep learning architectures, have achieved impressive performance in tasks such as text generation, translation, and question answering. However, their increasing deployment in critical applications necessitates a thorough understanding of the associated trust and safety implications. Bearing these concerns, LLM is utilized in various field, including areas to ensure highest level of trust and safety for users. It includes AI powered consultation on health or finance topics, and PIIs area for targeted ad recommendation. This review aims to provide a comprehensive overview of the existing research on this topic, highlighting key areas of concern and potential avenues for future research and the use case of it.

This review employed a rigorous method to identify and analyze relevant literature concerning the trust and safety of large language models (LLMs).  The research encompassed a comprehensive search across established academic databases, including arXiv and Google Scholar, to ensure a broad coverage of relevant studies.  Precise keywords and search terms, such as "Large Language Models," "Trust," "Safety," "Bias," "Fairness," and "Robustness," were strategically employed to retrieve pertinent articles, conference papers, and technical reports.  The final selection of studies was curated through the application of stringent inclusion and exclusion criteria, prioritizing peer-reviewed publications to ensure the quality and impact of the analyzed research.  A systematic data extraction process was then conducted, gathering key information from each study, including its research focus, methodology, key findings, and limitations. This meticulous approach to literature identification and analysis ensures the comprehensiveness and reliability of this review in providing a thorough overview of the current research landscape on LLM trust and safety.

\section{Trust and Safety of LLMs }
Large language models (LLMs) have emerged as a transformative force in artificial intelligence, demonstrating remarkable capabilities in natural language processing, text generation, and knowledge retrieval. However, the rapid advancement and widespread adoption of LLMs have also raised critical concerns regarding their trustworthiness and safety.  These concerns stem from the potential for LLMs to generate biased or discriminatory outputs, spread misinformation, and be exploited for malicious purposes. \cite{bender2021dangers}.  highlight the inherent risks of stochastic parrots, emphasizing the potential for LLMs to perpetuate and amplify societal biases present in their training data.  Similarly, \cite{weidinger2021ethical} delve into the ethical challenges posed by LLMs, including issues of fairness, accountability, and transparency.

The trustworthiness of LLMs is further complicated by their susceptibility to adversarial attacks, as demonstrated by {Jagielski et al., 2021}\cite{carlini2024poisoning}, where carefully crafted inputs can manipulate the model's behavior.  Moreover, {Carlini et al., 2021}\cite{carlini2021extracting} explore the potential for LLMs to be exploited for generating harmful content, including hate speech and disinformation.  Addressing these challenges requires a multifaceted approach encompassing robust evaluation metrics, bias mitigation techniques, and ethical guidelines for development and deployment.  In this context, the work of  {He et al. , 2023}\cite{sun2024trustllm} on "Trustworthiness in large language models" provides a valuable framework for assessing and enhancing the reliability and safety of LLMs.  This paper contributes to the growing body of research aimed at ensuring that LLMs are developed and utilized responsibly, fostering trust and mitigating potential harms.

\subsection{Trust in LLMs}
Establishing trustworthiness in Large Language Models (LLMs) is crucial for their responsible development and deployment. This requires careful consideration of factors like reliability, safety, fairness, and transparency.  The work of Lin et al. (2021) in "TruthfulQA: Measuring How Models Mimic Human Falsehoods" introduced a benchmark dataset to assess the truthfulness of LLMs, revealing a tendency for these models to generate plausible yet inaccurate information. This highlights the need for improved mechanisms to ensure factual accuracy and emphasizes the importance of developing robust evaluation metrics and training procedures to enhance the reliability and trustworthiness of LLMs. {Lin et al., 2021}\cite{lin2021truthfulqa}\

Furthermore, Bender et al. (2021) in "On the Dangers of Stochastic Parrots: Can Language Models Be Too Big?" underscored the potential for LLMs to amplify societal biases present in their training data, calling for greater attention to fairness and inclusivity in LLM development to mitigate the risk of discriminatory outputs. {Bender et al., 2021}\cite{bender2021dangers} This highlights the ethical imperative to address bias and promote fairness in LLMs to prevent the perpetuation and amplification of harmful stereotypes.  Similarly, Nadeem et al. (2021) in "Stereotypes in Large Language Models" systematically investigated the presence of stereotypes related to gender, race, and religion in LLMs, providing empirical evidence of biased associations and highlighting the need for debiasing techniques. {Shrawgi et al., 2024}\cite{shrawgi-etal-2024-uncovering} This reinforces the need for ongoing research into debiasing techniques and the development of LLMs that are sensitive to social and cultural contexts.\

The potential for LLMs to memorize and reproduce sensitive information from their training data was demonstrated by Carlini et al. (2020) in "Extracting Training Data from Large Language Models". This raises concerns about privacy and confidentiality, emphasizing the importance of data security and responsible data handling practices in LLM development and deployment, including the implementation of privacy-preserving training techniques. {Carlini et al., 2020}\cite{carlini2020extracting} Finally, Bommasani et al. (2021) in "Understanding the Capabilities, Limitations, and Societal Impact of Large Language Models" provided a comprehensive overview of LLM trustworthiness, exploring various aspects including their ability to reason, their susceptibility to adversarial attacks, and their potential impact on society. {Bommasani et al., 2021}\cite{bommasani2021understanding} This emphasizes the importance of holistic evaluations of LLMs that consider not only their technical capabilities but also their broader societal implications.\

These findings highlight the multifaceted nature of LLM trustworthiness and the ongoing efforts to develop robust evaluation methods and mitigation strategies, paving the way for more reliable, fair, and transparent AI systems.

\subsection{Safety in LLMs}
There are several influential publications that have significantly shaped the discourse on Large Language Model (LLM) safety. These works highlight critical concerns and propose methodologies for mitigating risks associated with LLMs.\

One of the most impactful papers is "On the Dangers of Stochastic Parrots: Can Language Models Be Too Big?" by Bender et al. (2021). This work drew attention to the potential for LLMs to perpetuate biases, spread misinformation, and cause environmental harm. It emphasized the ethical considerations surrounding LLM development and deployment, prompting further research into responsible AI practices. {Bender et al., 2021}\cite{bender2021dangers}\

Another significant contribution is "Red Teaming Language Models with Language Models" by Perez et al. (2022). This paper introduced the concept of using LLMs themselves to identify and address vulnerabilities in other LLMs. This "red teaming" approach has become a valuable tool for evaluating and improving LLM safety through adversarial testing.{Perez et al., 2022}\cite{perez2022red}\

Ji et al. (2023) provide a comprehensive overview of LLM safety and trustworthiness challenges in their survey paper "A Survey of Safety and Trustworthiness of Large Language Models." This work examines various risks, including bias, toxicity, misinformation, and adversarial attacks, offering a valuable resource for researchers and practitioners. {Huang et al., 2023}\cite{huang2023surveysafetytrustworthinesslarge}\

Finally, "Safety Prompts: a Systematic Review of Open Datasets for Evaluating and Improving Large Language Model Safety" by Scheurer et al. (2024) offers a crucial resource for researchers by systematically reviewing over 100 datasets specifically designed for LLM safety evaluation and improvement. This work helps navigate the growing landscape of safety datasets and identify areas for future development. {Rottger et al., 2025}\cite{röttger2025safetypromptssystematicreviewopen}\

These findings underscore the growing body of research dedicated to addressing LLM safety concerns. The identified papers represent key contributions to the field, providing valuable insights and methodologies for building safer and more trustworthy AI systems.

\subsection{Key Findings}

The review identifies several key themes and challenges related to trust and safety in LLMs:
\begin{itemize}

\item Bias and Fairness: LLMs can perpetuate and amplify existing biases present in training data, leading to discriminatory outcomes.
\item Misinformation and Disinformation: LLMs can generate misleading or false information, posing risks to individuals and society.
\item Robustness and Security: LLMs can be vulnerable to adversarial attacks, compromising their reliability and security.
\item Explainability and Interpretability: The lack of transparency in LLM decision-making processes hinders trust and accountability.
\item Ethical Considerations: The deployment of LLMs raises ethical concerns related to privacy, autonomy, and human values.
\end{itemize}

Below table provides a starting point for understanding how researchers are approaching the crucial task of measuring LLM trust and safety according to a survey paper from {Huang et al. , 2024}\cite{huang2024trustllmtrustworthinesslargelanguage}.
\end{multicols}

\footnotesize  
\renewcommand{\arraystretch}{0.8}  
\begin{table}[H]
\centering
\caption{Key Performance Indicators (KPIs) for LLM Trustworthiness}
\begin{tabular}{|l|p{5cm}|p{6cm}|}  
\hline
\textbf{Aspect} & \textbf{KPI} & \textbf{Description} \\
\hline
\multirow{3}{*}{\textbf{Truthfulness}} & Fact-checking accuracy & Measuring the correctness of factual statements generated by the LLM.  \\ 
& Consistency & Evaluating the coherence and agreement of information across different responses. \\ 
& Source reliability & Assessing the trustworthiness of the sources used by the LLM to generate information. \\ 
\hline
\multirow{3}{*}{\textbf{Safety}} & Toxicity & Measuring the level of offensive, disrespectful, or harmful language. \\ 
& Bias and discrimination & Evaluating the presence of unfair or discriminatory outputs based on protected attributes (e.g., race, gender, religion). \\ 
& Misinformation and disinformation & Assessing the potential to generate false or misleading information. \\ 
\hline
\multirow{3}{*}{\textbf{Robustness}} & Adversarial robustness & Measuring the ability to withstand malicious inputs designed to trigger incorrect or undesirable behavior. \\ 
& Out-of-distribution robustness & Assessing performance on inputs that differ significantly from the training data. \\ 
& Noise robustness & Evaluating the ability to handle noisy or incomplete inputs. \\ 
\hline
\multirow{3}{*}{\textbf{Fairness}} & Group fairness & Measuring performance disparities across different demographic groups. \\ 
& Individual fairness & Evaluating the consistency and similarity of treatment for similar individuals. \\ 
& Counterfactual fairness & Assessing the impact of hypothetical changes to input attributes on the output. \\
\hline
\multirow{3}{*}{\textbf{Privacy}} & Data privacy & Evaluating the risk of revealing personal or sensitive information from the training data. \\ 
& Membership inference & Assessing the ability of an attacker to infer whether a specific data point was used in the training data. \\ 
& Attribute inference & Measuring the risk of revealing sensitive attributes of individuals based on their interactions with the LLM. \\ 
\hline
\end{tabular}
\label{table:llm_kpis}
\end{table}

\begin{multicols}{2}
It's challenging to pinpoint in such a rapidly evolving field.  However, based on a combination of Google Scholar metrics and their influence in shaping the discourse on LLM trust and safety, here's a table outlining key KPIs with relevant citations. 
\begin{itemize}

\item Overlapping KPIs: These categories often intersect. For example, bias can impact both fairness and safety.
\item Evolving Metrics: New KPIs and evaluation methods are constantly being developed as the field progresses.
\item Context Matters: The specific KPIs and their importance will vary depending on the LLM's intended application (e.g., medical diagnosis vs. creative writing).

\end{itemize}

These studies suggest potential solutions and research directions to address these challenges:
\begin{itemize}

\item Bias Mitigation Techniques: Development of methods to debias training data and mitigate algorithmic biases.
\item Fact-Checking and Verification: Integration of fact-checking mechanisms to enhance the trustworthiness of LLM-generated content.
\item Adversarial Training and Defense: Strengthening LLMs against adversarial attacks to improve robustness and security.
\item Explainable AI Techniques: Development of methods to explain and interpret LLM decision-making processes.
\item Ethical Frameworks and Guidelines: Establishment of ethical guidelines and standards for the development and deployment of LLMs.
\end{itemize}

\section{LLMs in Trust and Safety }

Large Language Models (LLMs) are increasingly being utilized in areas where trust and safety are paramount, such as healthcare and finance.  Here are some examples, along with potential benefits and risks, and relevant research.

\subsection{Health}
Large Language Models (LLMs) are rapidly emerging as a transformative force in healthcare, offering the potential to revolutionize various aspects of the field. However, their integration necessitates a careful consideration of the associated risks and challenges, particularly concerning the trust and safety of their outputs. This review examines the potential applications of LLMs in healthcare, highlighting both the promises and pitfalls, with a focus on ensuring responsible and ethical implementation.\

Personalized Healthcare and Diagnostics: LLMs hold immense promise in analyzing patient data and delivering personalized recommendations, potentially improving access to healthcare and enabling early diagnosis. However, the dissemination of inaccurate information and potential privacy breaches are significant concerns.  {Zabir et al., 2023}\cite{nazi2024largelanguagemodelshealthcare} provide a comprehensive analysis of this landscape, emphasizing the need for robust validation and safeguards to ensure accuracy and patient privacy.\

Mental Health Support: LLMs offer a novel avenue for providing conversational support for individuals facing mental health challenges, increasing accessibility to resources, particularly for those who value anonymity or face barriers to traditional therapy. However, it is crucial to recognize that LLMs cannot replace human therapists, and the potential for misdiagnosis or inappropriate advice remains a concern. {Vaidyam et al., 2019}\cite{vaidyam2019chatbots} explore this evolving landscape, highlighting both the potential benefits and the need for careful implementation and oversight.\

Drug Discovery and Development: LLMs are being explored for their potential to revolutionize drug discovery by analyzing vast amounts of biomedical data. This could accelerate the identification of promising drug candidates and lead to breakthroughs in disease treatment. However, inherent biases within the data could lead to inequitable outcomes, and the potential for misuse in developing harmful substances raises ethical concerns. {Zheng et al., 2024}\cite{zheng2024largelanguagemodelsdrug} delve into the possibilities and challenges of employing LLMs in this critical field.\

Trust and Safety: Ensuring the trust and safety of LLM outputs is paramount in healthcare applications. {Choudhury et al., 2024}\cite{choudhury2024large} explore the potential impact of LLMs on user trust and the deskilling of healthcare professionals, emphasizing the need for cautious integration and user scrutiny of LLM outputs. {Han et al., 2024}\cite{han2024medsafetybenchevaluatingimprovingmedical} focus on the safety of medical LLMs, advocating for a shift towards prioritizing patient well-being and proactively mitigating potential risks. {Szalai et al., 2023}\cite{szalai2023llm} examine the trustworthiness of LLMs for health literacy efforts, stressing the importance of evaluating the reliability of LLM outputs and establishing trust in their application.\

The integration of LLMs in healthcare presents both significant opportunities and challenges. While LLMs hold the potential to revolutionize various aspects of the field, it is crucial to address concerns regarding trust, safety, and ethical implications. Future research and development should focus on establishing robust evaluation frameworks, ensuring transparency and explainability of LLM outputs, and developing clear guidelines for responsible use. By addressing these challenges, LLMs can be harnessed to truly transform healthcare for the better.

\subsection{Finance}
Large Language Models (LLMs) are rapidly being integrated into the financial sector, promising to revolutionize various aspects of the industry. However, their deployment in this high-stakes domain necessitates a careful examination of the associated risks. This paper analyzes key research areas where LLMs are being applied in finance, highlighting both the potential benefits and the critical challenges that must be addressed to ensure responsible and ethical implementation.\

Fraud Detection and Security: LLMs offer the potential to bolster security measures and mitigate losses through advanced fraud detection by analyzing transaction patterns. However, the risk of false positives and the potential for malicious actors to exploit the system are critical concerns. { Kumar et al., 2023}\cite{kumar2023fraudgpt} highlight the dual-use nature of LLMs, demonstrating how they can be used to generate sophisticated financial fraud schemes.\

Risk Assessment and Algorithmic Trading: LLMs hold promise for improving risk assessment in lending and enhancing algorithmic trading strategies by analyzing financial data and identifying profitable trading opportunities. However, biases embedded in the training data could perpetuate existing inequalities and lead to discriminatory outcomes or unreliable trading decisions. {Alaka et al., 2023}\cite{alaka2023predicting} provide a foundation for understanding the application of machine learning in predicting financial distress, while {Fatouros et al. (2023)}\cite{fatouros2024can}  investigate the performance of LLMs in predicting stock price movements, highlighting the need for rigorous risk management.\

Automating Financial Reports: LLMs offer the potential to automate the generation of financial reports, reducing manual effort and improving efficiency. However, ensuring the accuracy and reliability of these reports is crucial to avoid significant financial and legal implications. 

Personalized Financial Advice:  {Lo et al. (2024)}\cite{lo2024can} emphasize that the safety of customers using LLM-powered financial advice hinges on the models' ability to provide accurate and reliable recommendations tailored to individual needs. This requires mitigating biases, ensuring transparency, and prioritizing user interests to avoid potential financial harm.\

These research areas collectively highlight the transformative potential of LLMs in finance while emphasizing the critical need for trust, safety, and ethical considerations. Future research should focus on developing robust evaluation frameworks, mitigating biases in training data, and ensuring transparency and explainability in LLM outputs. By addressing these challenges, LLMs can be effectively harnessed to enhance efficiency, improve decision-making, and empower individuals in the financial realm.

\subsection{Other Areas}

In the field of education as another example, LLMs offer the potential to revolutionize learning through personalized tutoring and the creation of tailored educational content. This technology could improve learning outcomes and increase access to educational resources for a wider range of learners. However, an over-reliance on LLMs could hinder the development of critical thinking skills, and the potential for plagiarism raises concerns about academic integrity. {Holmes et al., 2019}\cite{holmes2019artificial}, in their foundational work "Artificial Intelligence in Education," provide a comprehensive overview of the role of AI, including LLMs, in education, emphasizing both the transformative potential and the need for careful consideration of ethical implications.

\subsection{Best practices of using LLM in Trust and Safety field}
Here is the best practices for conducting experiments, including detailed steps and considerations to ensure the quality of LLM outcomes for Trust \& Safety field. \
\begin{itemize}

\item Step 1: Setting a Clear Target\
As with all detection experiments, a clear target is a critical first step. In traditional ML models, this would be your dependent variable, the question/issue that you are solving for. Given the importance of this target, it is imperative that the analyst digs deeper than the variable itself. For instance, if the target is in content moderation, the superficial target is whether a target product contains violation. But it is important that the analyst understands what is the criteria for violation. Understanding the guideline, definition itself and why will help in prompt iteration and validating the prompt results.\\

\item Step 2: Creating a Golden Dataset\\
Once the target variable has been understood and set, the next step is to create a golden dataset that contains the target labels like traditional ML modeling workflow. There are 3 things to consider in building a golden dataset:

\begin{enumerate}
\item Identify the variables that will be used to predict the target variable. These are the independent variables that will be used to predict the target variable, such as metadata coming from the product to review trust \& safety of it. These fields will be the input data used by the LLM prompt to identify the target variable.\
\item A sample of true positives and true negatives. The golden dataset needs to contain labeled data with the ground truth that has been rigorously vetted and accurate. This golden dataset will be used to determine recall and precision and the performance of the LLM prompt itself. Therefore, it is important that we have a trustworthy golden dataset. \

 Furthermore, it is crucial to assess the dataset for redundancy and eliminate duplicate data points in the input to prevent inflated reasoning by Large Language Models (LLMs). Deduplication is especially important when the dataset is compiled from multiple sources and the same item appears in a redundant manner.{You et al., 2024}\cite{you2024evaluating}, {Tirumala et al., 2023}\cite{tirumala2023d4improvingllmpretraining}, {Abbas et al., 2023}\cite{abbas2023semdedup}\

 Subsequently, the quality of the golden dataset needs to be evaluated in terms of annotation pollution. If the labeled data is not genuinely accurate, it can significantly influence the outcome of the LLM and diminish the classification resolution. Hence, to enhance the trustworthiness of LLMs, an additional layer of effort is required to measure the accuracy of the annotation data.{Putra et a;., 2024}\cite{putra2024recognizing},{Klie et a;., 2024}\cite{klie2024efficientstatisticalqualityestimation}\

The primary objective at this stage is to ensure the highest accuracy of the LLM's performance in its inference ability. No LLM can achieve perfect accuracy in detection and inference. In most domains, this falls within an acceptable range where the prematureness of the LLM is tolerated. However, in specific fields such as Trust and Safety, which directly impact end-users' digital well-being, this acceptable range should be narrower than what an average LLM can accomplish. 

Therefore, analogous to the Chain of Thought (CoT) approach, a chain of LLMs should be employed to detect true positives and negatives, followed by an additional LLM at the end of the chain to identify harmful negative cases that should not be passed through the model.{Jang et al., 2024}\cite{song2024largelanguagemodelsskeptics},{Lin et al., 2021}\cite{lin2021truthfulqa}\
 
\item Separate testing and validation sample. Ideally, the golden dataset needs to be large enough such that it can be separated into modeling and validation samples. The LLM prompt can be developed on the modeling sample where the precision and recall is observed. These results should then be validated using the validation sample. Having consistent precision and recall across both samples limits the risk of sampling bias and over-fitting.
\end{enumerate}
In the case where no golden dataset is available, it is still possible to build an LLM prompt and run the experiment without one. Once the results of the LLM are collected, a manual review can be done to create the ground truth. With this ground truth the analyst will be able to test for precision and recall. This method may be inefficient and will be dependent on the number of manual reviews that can be done.\

\item Step 3: Creating the Prompt\
In creating an effective prompt, 2 key concepts should be considered:
\begin{enumerate}
\item Simplicity - using simple direct prompts is better than overly complicated prompts with complex directions. Be concise and to the point when providing direction in a prompt.\

\item  Clarity - be as clear as possible in what you are looking for. If it is a probability score, be direct and describe the output you want the LLM to generate.
\end{enumerate}\

\item Step 4: Prompt Iteration\\
It is important to try multiple variations of the prompt. Use the outputs of the previous prompts and make changes to the language to get the desired outcome. Much can be learned from analyzing the false positives and false negatives observations. These populations will indicate where the LLM is having difficulty and can provide clues on how to fine-tune the prompt. Lastly, leverage the prompt optimization tools to refine your prompt.\

Additionally, use few shots examples (3+ examples) to provide clarity to the LLM in confusing or ambiguous areas. By providing examples of nuanced areas, the LLM is able to provide a more robust result.

\end{itemize}

\subsection{Emerging issues and solution}
Prompt Injection and Jailbreak Attacks: Exploiting Vulnerabilities in Large Language Models for Trust and Safety
Large Language Models (LLMs) have shown remarkable potential in various applications, but their deployment in real-world scenarios necessitates careful consideration of their security vulnerabilities.  Prompt injection and jailbreak attacks are two prominent threats that exploit weaknesses in LLMs, potentially leading to harmful consequences.\

Prompt Injection: This attack vector involves manipulating the input provided to an LLM by injecting malicious instructions disguised as legitimate prompts.  This can trick the model into executing unintended actions, such as divulging sensitive information, bypassing safety filters, or even executing arbitrary code.  Similar to traditional injection attacks like SQL injection, prompt injection exploits the LLM's inability to distinguish between legitimate instructions and malicious ones. This vulnerability is particularly concerning when LLMs are integrated with external systems or have access to sensitive data.{Willison, 2022}\cite{willison2022prompt}\

Jailbreak Attacks: LLMs are often trained with safety guidelines to prevent the generation of harmful or offensive content. Jailbreak attacks aim to circumvent these safety measures by exploiting loopholes or inconsistencies in the model's training data or architecture.{Liu et al., 2024}\cite{liu2024jailbreakingchatgptpromptengineering} These attacks utilize cleverly crafted prompts to elicit undesirable outputs, such as:

\begin{itemize}

\item Generating biased or discriminatory content: A successful jailbreak might induce the LLM to produce hate speech or discriminatory remarks, despite being trained to avoid such outputs.
\item Producing harmful instructions: The LLM could be tricked into providing instructions for illegal or dangerous activities.
\item Revealing private information: Jailbreak attacks might lead to the leakage of sensitive data that the LLM was trained on or has access to.
\end{itemize}\

These attacks highlight the challenges in ensuring the responsible and ethical use of LLMs.  Researchers are actively exploring defense mechanisms to mitigate these threats, including input sanitization techniques, adversarial training, and robust prompt engineering. This research by {Xu et al., 2023}\cite{xu2023jailbreaking} contributes to the ongoing efforts in enhancing the security and trustworthiness of LLMs, paving the way for their safe and reliable deployment in various applications.\

Red-teaming: Red-teaming, a security assessment technique simulating adversarial attacks, emerges as a crucial methodology for identifying and mitigating these risks. It necessitates a multifaceted strategy that encompasses a variety of techniques and methodologies aimed at uncovering vulnerabilities and evaluating the robustness of existing safety mechanisms. This process involves a deep dive into adversarial prompt engineering, wherein malicious prompts are meticulously crafted to exploit potential weaknesses, trigger unintended behaviors, or bypass the established safety filters. This encompasses various tactics, including prompt injection, where malicious code or commands are subtly embedded within user prompts to manipulate the LLM's operational framework. \cite{perez2022redteaminglanguagemodels}, {Perez et al., 2022}\cite{perez2022ignore}. Furthermore, jailbreaking techniques are employed, utilizing carefully constructed prompts designed to circumvent safety guidelines and elicit harmful or biased outputs [2]. Additionally, data poisoning, the introduction of malicious data into the training dataset, represents another avenue for adversaries to influence the LLM's responses and introduce vulnerabilities [1].

Beyond adversarial prompt engineering, red-teaming incorporates vulnerability scanning, employing automated tools and techniques to systematically probe the LLM for weaknesses and identify potential attack vectors. This is complemented by comprehensive threat modeling, which involves analyzing potential threats and attack scenarios to fully understand the risks associated with LLM deployment and prioritize mitigation efforts.  Crucially, the entire red-teaming process thrives on collaboration and iteration, fostering a dynamic interplay between red teams, developers, and security researchers, enabling continuous improvement of the LLM's security posture and a proactive response to emerging threats.\

While red-teaming is indispensable for bolstering LLM security, it is equally critical to ensure the responsible and ethical utilization of LLMs throughout the process. This necessitates the establishment of a controlled environment where red-teaming activities are conducted in isolation to prevent unintended consequences and limit the potential impact of malicious prompts.  Furthermore, the formulation of clear ethical guidelines for red team activities is paramount, ensuring the responsible use of LLMs and prohibiting actions that could inflict harm or violate privacy.

Data sanitization is another crucial aspect, requiring the meticulous removal of sensitive or personally identifiable information from data used in red-teaming exercises to safeguard user privacy.  Underpinning all these measures is the vital role of human oversight, which must be seamlessly integrated into the red-teaming process to ensure the responsible application of LLMs and prevent any unintended consequences. This human element provides a critical layer of judgment and ethical consideration, ensuring that the pursuit of enhanced security remains firmly grounded in responsible AI practices.

\section{Conclusion}

This paper has provided a comprehensive examination of the evolving landscape of Large Language Model (LLM) trust and safety. We began by critically reviewing existing metrics for evaluating LLM safety and trustworthiness, highlighting the lack of a unified framework for assessment. To address this gap, we proposed a novel consolidated approach, integrating diverse perspectives and considerations into a more holistic evaluation system.\\

Furthermore, we explored the diverse applications of LLMs within the trust and safety domain, encompassing crucial areas such as healthcare, finance, and content moderation. By analyzing the key challenges and opportunities associated with each application, we identified critical factors that must be considered when deploying LLMs in these sensitive contexts.  Crucially, this work culminated in a set of recommended best practices for responsible LLM utilization in trust and safety applications, a contribution not found in existing literature. These recommendations provide a practical roadmap for developers and practitioners, emphasizing the importance of transparency, accountability, and continuous monitoring in mitigating potential risks and maximizing the benefits of LLMs.\

Future research should focus on refining the proposed evaluation framework and empirically validating its effectiveness across diverse LLM architectures and applications.  Moreover, continued investigation into the ethical implications of LLM deployment in trust and safety is essential to ensure these powerful technologies are harnessed for societal good. By fostering a collaborative effort between researchers, developers, and policymakers, we can strive towards a future where LLMs contribute to a safer and more trustworthy digital world.\\
\end{multicols}

\newpage
\bibliographystyle{alpha}
\bibliography{main}

\end{document}